\documentclass[runningheads]{llncs}

\usepackage[T1]{fontenc}
\usepackage{graphicx,verbatim}
\usepackage{cite}
\usepackage{amsmath}
\usepackage{color}
\usepackage{multirow}
\usepackage{booktabs}
\usepackage{amssymb} 
\usepackage{array} 
\usepackage{makecell} 
\usepackage{float}
\usepackage{subcaption} 
\usepackage{fontawesome} 
\usepackage[misc]{ifsym}

\begin{document}

\title{Omni-Fusion of Spatial and Spectral for Hyperspectral Image Segmentation}
\author{Qing Zhang \and Guoquan Pei \and Yan Wang\textsuperscript{(\Letter)}}
\institute{Shanghai Key Laboratory of Multidimensional Information Processing, East China Normal University, Shanghai 200241, China\\
\email{ywang@cee.ecnu.edu.cn}}

\maketitle              

\begin{abstract}
Medical Hyperspectral Imaging (MHSI) has emerged as a promising tool for enhanced disease diagnosis, particularly in computational pathology, offering rich spectral information that aids in identifying subtle biochemical properties of tissues. Despite these advantages, effectively fusing both spatial-dimensional and spectral-dimensional information from MHSIs remains challenging due to its high dimensionality and spectral redundancy inherent characteristics. To solve the above challenges, we propose a novel spatial-spectral omni-fusion network for hyperspectral image segmentation, named as Omni-Fuse. Here, we introduce abundant cross-dimensional feature fusion operations, including (1) a cross-dimensional enhancement module that refines both spatial and spectral features through bidirectional attention mechanisms; (2) a spectral-guided spatial query selection to select the most spectral-related spatial feature as the query; and (3) a two-stage cross-dimensional decoder which dynamically guide the model’s attention towards the selected spatial query. Despite of numerous attention blocks, Omni-Fuse remains efficient in execution. Experiments on two microscopic hyperspectral image datasets show that our approach can significantly improve the segmentation performance compared with the state-of-the-art methods, with over 5.73\% improvement in DSC. Code available at: https://github.com/DeepMed-Lab-ECNU/Omni-Fuse.
\end{abstract}
\renewcommand{\thefootnote}{}
\footnotetext{Q. Zhang and G. Pei—Denotes equal contribution.}

\section{Introduction}
\begin{figure}[t]
    \centering
    \includegraphics[width=\linewidth]{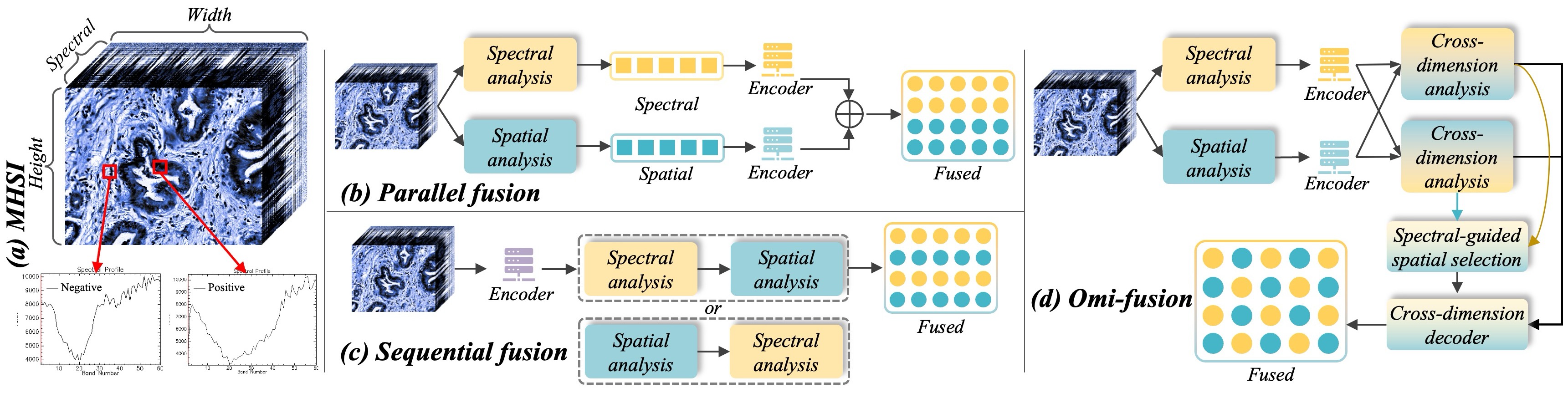}
    \caption{(a) Microscopic hyperspectral image (MHSI) of Multi-Dimensional Choledoch Dataset (MDC) dataset. Positive and Negative regions have different spectral curves and single band images at various wavelengths can highlight different texture information. (b)-(d) Structures for spatial-spectral feature fusion.}
    \label{fig:hsi_intro}
\end{figure}

Hyperspectral Imaging (HSI) is an emerging technology in pathological diagnosis, reflecting the physiological and biochemical properties of different tissue components of tissues in a non-invasice manner~\cite{bengs2020spectral,banu2024hyperspectral}.
As shown in Fig.~\ref{fig:hsi_intro} (a), each single-band image in a microscopic hyperspectral image (MHSI) represents the reflectance spectrum information within a specific wavelength range, with different wavelengths highlighting various pathological regions.
Therefore, each pixel has a corresponding spectral curve, ensuring that the granularity of both spatial and spectral features is aligned~\cite{li2022asymmetric,zhang2023hyperspectral}.
Mining the correlations and complementary information between spectral and spatial data enhance the model's ability to understand and analyze complex pathological scenes.

MHSIs contain asymmetric information in spectral and spatial dimensions. The key challenge for MHSI segmentation is how to effectively align this cross-dimensional information. Recent attempts either learn spectral and spatial features in a sequential or a parallel manner \cite{dong2024multi}, and try to fuse features in the encoder. Since the former explicitly processes one dimension at each stage \cite{yun2023spectral,xie2023exploring,yao2024specat}, it is hard to enable the two dimensional features to simultaneously interact with each other. The latter always designs a dual stream to extract spectral and spatial features, respectively, and then followed by post-fusion, either in the late stage of the encoder or in the decoder \cite{yun2023factor,song2024interactive}. Spectral and spatial features do not fully guide each other in this way, leading to the loss of complementary information. Though FL \cite{dong2024multi} integrates spatial and spectral features for a spatiospectral balance, they simply apply deformable cross-attention twice between the encoder and the decoder. We summarize the network architecture design in Fig.~\ref{fig:hsi_intro} (b) and (c). It shows that none of the above networks provides abundant multi-dimensional feature fusion, thereby leading to sub-optimal results.

\begin{figure}[t]
    \centering
    \includegraphics[width=0.95\linewidth]{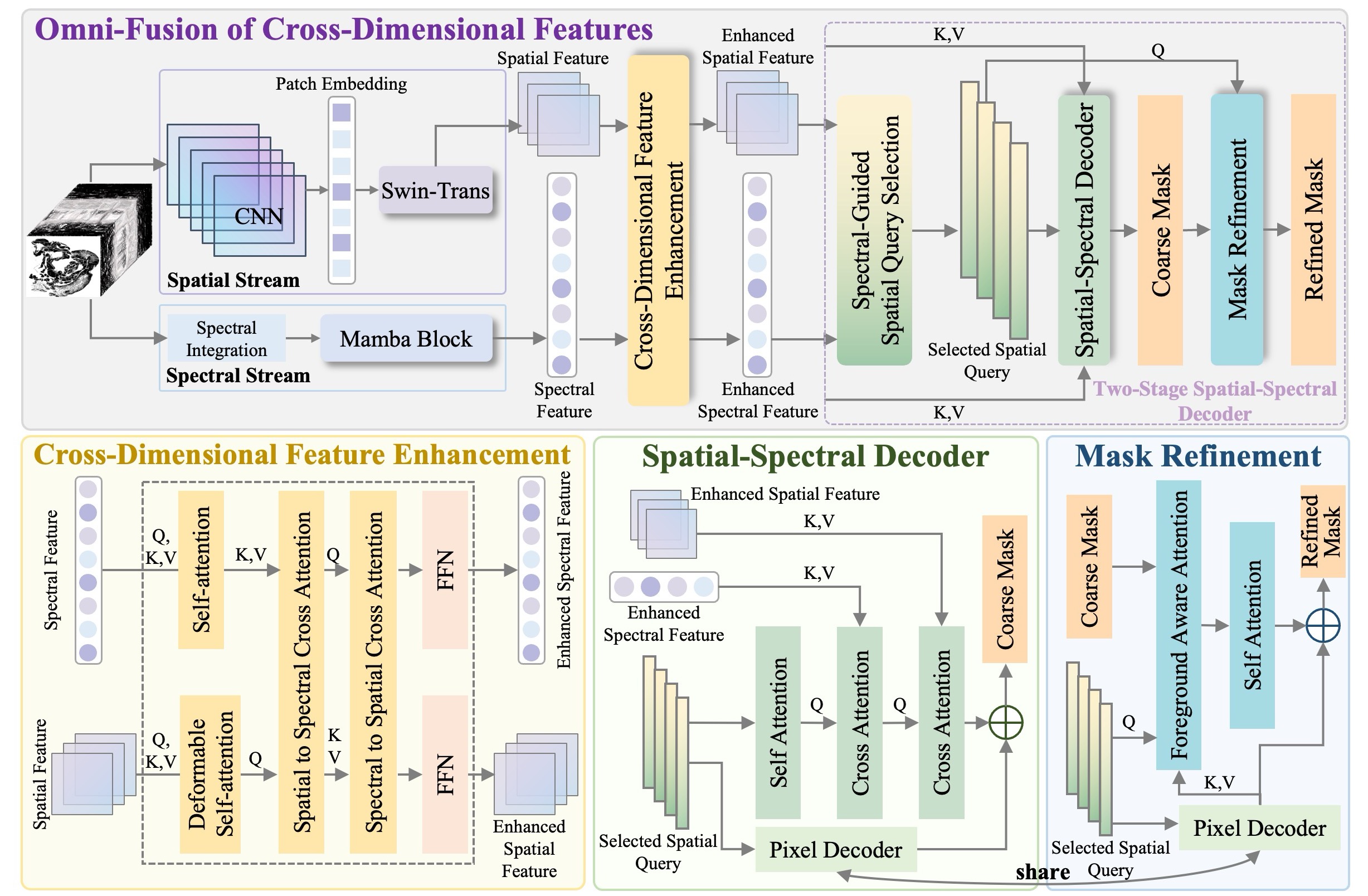}
    \caption{Framework architecture of Omni-Fuse. The MHSI is primarily processed by a dual steam, followed by a cross-dimensional attention based feature enhancer. And the enhanced spatial features are further compacted by a spectral-guided spatial selection. Subsequently, the compacted features are decoded cooperate with enhanced spatial and spectral features to generate a coarse mask. Finally, the coarse mask is further refined by mask attention mechanism.}
    \label{fig:framework}
\end{figure}

In this paper, we are the first to achieve abundant multi-dimensional feature fusion for MHSI segmentation.
Similar with parallel fusion, we separately extract primary spectral and spatial features by a dual stream. Afterwards, we design our spatial-spectral omni-fusion, which makes fusion omnipresent. Our fusion lies in the \textbf{feature enhancer} ( ``Cross-dimensional analysis'' in Fig.~\ref{fig:hsi_intro}), the \textbf{query selection} (``Spectral-guided spatial selection'' in Fig.~\ref{fig:hsi_intro}) and the \textbf{decoder} (``Cross-dimension decoder'' in Fig.~\ref{fig:hsi_intro}). 
For feature enhancer, the primary features mutually enhance each other through two designed bidirectional cross-dimensional analysis blocks, extracting the most important and relevant spatial-spectral feature (see ``Cross-Dimensional Feature Enhancement'' in Fig.~\ref{fig:framework}).
The enhanced spatial features are further compacted according to their correlation with the enhanced spectral feature (see ``Spectral-Guided Spatial Query Selection'' in Fig.~\ref{fig:framework}).
For decoder, we design a coarse-to-fine learning strategy to extract fine-grained localized features.
Specifically, we propose a spectral-spatial decoder for coarse segmentation, using enhanced spectral and spatial features to dynamically guide the model's attention towards the selected spatial query (see ``Spatial-Spectral Decoder'' in Fig.~\ref{fig:framework}).
Then, we refine the coarse mask by incorporating with the selected spatial query (see ``Mask Refinement'' in Fig.~\ref{fig:framework}).
Through iterative interactions, spatial and spectral features are deeply and fully fused in this paper.
Our contributions are as follows:
\begin{itemize}
    \item We propose an extensive Omni feature fusion network (Omni-Fuse) for microscopic hyperspectral image segmentation, introducing a cross-dimensional enhancement to bidirectionally integrate spatial-spectral features.
    \item We design a two-stage spatial-spectral decoder to coarsely-then-finely segment tumor regions with the assistance of SAM decoder.
    \item Comprehensive experiments are conducted on a public MDC Dataset and a private GPCC Dataset, proving the effectiveness and efficiency of the proposed multi-modal feature fusion based segmentation method.
\end{itemize}

\section{Method}
\subsection{Problem Definition and Network Structure}
Given a pathological microscopic hyperspectral image dataset $\mathcal{D}=\{\mathbf{X},\mathbf{Y}\}$ where $\mathbf{X} \in \mathbb{R}^{H \times W \times S}$ represents the image with spatial resolution of $H \times W$ and spectral band number of $S$, our goal is to predict its pixel-level segmentation map of $H \times W$.
Fig.~\ref{fig:framework} illustrates the architecture of the proposed abundant \textbf{Omni} feature \textbf{Fus}ion {N}etwork (\textbf{Omni-Fuse}) for MHSI segmentation, integrating a cross-dimension feature enhancement module and a two-stage spatial-spectral coarse-to-fine decoder, to predict pixel-level label $\hat{\mathbf{Y}}=\{0,1\}\in\mathbb{R}^{H\times W}$.
We first extract primary features through a simple dual stream, followed by a novel cross-dimensional feature enhancement module to extract the most important and relevant spatial-spectral features.
And the spatial feature is compacted by a spectral-guided spatial feature selection block.
Finally, a two-stage spatial-spectral decoder is designed for prediction generation from coarse to fine.

\subsection{Primary Spatial-Spectral Feature Extraction}
Due to the fact that microscopic hyperspectral image is a three-dimensional data cube, its data size is significantly larger compared to traditional RGB or grayscale image.
In hyperspectral pathology analysis tasks, how to efficiently extract and fuse these multi-dimensional features is a key challenge.
To address this issue, we adopt a dual-branch architecture to separately extract spatial information and spectral information.
For spatial feature extraction, we apply a Convolutional Neural Network (CNN) to capture local spatial feature $F_{\text{CNN}} \in \mathbb{R}^{H' \times W' \times C}$, where $C$ is the channel number, followed by Swin-Transformer encoder to capture long-range dependencies in the spatial domain, refining the spatial feature to $F_{\text{swin}} \in \mathbb{R}^{H' \times W' \times C}$.
For Spectral feature extraction, inspired by MDN~\cite{lin2025mdn}, we employ a bi-directional mamba based block specifically designed for capturing long-distance dependencies across the spectral domain, exporting $X_{\text{spec}} \in \mathbb{R}^{B \times H \times W \times S \times L_{\text{spec}}}$, where $B$ is the batch size and $L_{\text{spec}}$ is the dimension of the spectral features.
For the following feature fusion of spatial and spectral information, the spectral feature is flattened to continuous spectral tokens $T_{prispec} = \{[T_0, T_1, ..., T_s] \mid T_j \in \mathbb{R}^{L_{spec}}\}$.

\subsection{Cross-Dimensional Feature Enhancement}
Primary spatial and spectral features are extracted separately using simple CNN or Mamba-based algorithms, the three-dimensional information of the hyperspectral image remains inadequately analysed.
Therefore, we introduce a cross-dimensional enhancement module based on bidirectional attention mechanisms to integrate the aforementioned features.
Each pixel in the spatial dimension has its own spectral curve, consistent with the curve of the same category area and differing from the curve of areas belonging to different categories.
Since then, the granularity of spatial and spectral features is considered to be aligned.
To fully utilize the complementary information between spectral features and the structural information of spatial features, we propose a cross-dimensional feature enhancement module.
For spatial feature, considering that tissues and lesions in histopathological images often exhibit irregular and deformed shapes, deformable self-attention module is selected in this module to flexibly capture local details while efficiently modeling long-range morphological correlations, exporting updated spatial feature $F'_{spa} = \texttt{DeformableSelfAttn}(F_{swin})$.
We get updated spectral feature $F'_{spec}$ through regular self-attention mechanism to capture dependencies within the spectral domain.
Then, two-layer bidirectional cross-attention module is designed here to deeply achieve cross-dimensional communication:
\begin{equation}
    F''_{spa}=\texttt{CrossAttn}({F'_{spa}},{F'_{spec}})=\texttt{softmax}(\frac{F'_{spa}{F'_{spec}}^{\mathsf{T}}}{\sqrt{d_{spec}}})F'_{spec},
\end{equation}
\begin{equation}
F''_{spec}=\texttt{CrossAttn}({F'_{spec}}, {F'_{spa}})=\texttt{softmax}(\frac{F'_{spec}{F'_{spa}}^{\mathsf{T}}}{\sqrt{d_{spa}}})F'_{spa},
\end{equation}
where $\{F''_{spa},F''_{spec}\}$ are the enhanced features and $\texttt{CrossAttn}(\cdot)$ represents the vanilla cross attention operation. $\{d_{spa},d_{spec}\}$ are the dimension number of $\{F'_{spa},F'_{spec}\}$ respectively.
The extensively fused features greatly enhance the compatibility of feature expression in hyperspectral data. This enhancer not only makes the learned visual features spectral-aware and vice versa, but also provides spectral prompts for downstream segmentation tasks.

Usually, positive regions occupy only a small portion in pathological images, indicating that redundant information exists in the spatial features.
Therefore, we introduce a spatial feature selection module, exploring features that are strongly correlated with spectral information, for the following two-stage decoder.
For feature selection, we convert the enhanced features $\{F''_{spa},F''_{spec}\}$ into tokens through patch embedding, \emph{i.e.}, spatial token $T_{spa} \in \mathbb{R}^{N_{spa} \times d}$ and spectral token $T_{spec} \in \mathbb{R}^{N_{spec} \times d}$. Here, $N_{spa}$ is the number of spatial tokens, $N_{spec}$ is the number of spectral tokens, and $d$ is the number of feature dimension.
We select the most relevant $N_q$ spatial tokens $T'_{spa}$ by calculating the correlation between spatial and spectral tokens: 
\begin{equation}
    T'_{spa}=\texttt{Top}_{N_q}({\max}^{(-1)}(T_{spa} T_{spec}^\mathsf{T})),
\end{equation}
where $\texttt{Top}_{N_q}(\cdot)$ is the operation to select the most relevant $N_q$ spatial tokens. ${\max}^{(-1)}$ means the max operation along the -1 dimension.

\subsection{Two-Stage Spatial-Spectral Decoder}
To precisely segment tumor regions from the microscopic hyperspectral image, we design a two-stage cross-dimension decoder to further combine enhanced spatial and spectral features, \emph{i.e.}, the spatial-spectral decoder and the mask refinement. 
In the first stage, we combine attention-based cross-dimension decoder and SAM's pixel decoder~\cite{kirillov2023segment} to create a coarse mask.
Specifically, the spatial query $T'_{spa}$ is fed into a self-attention layer, followed by a spatial cross-attention layer and a spectral cross-attention layer to fuse the enhanced features $\{F''_{spa},F''_{spec}\}$.
We then apply a linear layer for the combination of deeply fused features and pixel decoder's output feature, to generate a coarse mask $P_c\in \mathbb{R}^{H/4 \times W/4}$.

To refine the segmentation, in the second stage, we follow Mask2Former~\cite{cheng2021maskformer} to enhance the object details with a foreground aware attention operation as calculated in Eq.~\ref{eq:maskattention}.
Through multiplying the mask with the attention map, the model can focus more on foreground regions and prove the tumor boundary:
\begin{equation}
    F_{o} = \texttt{SelfAttn}(\texttt{FgroundAttn}(T'_{spa},F_{pd},P_c)),
\label{eq:maskattention}
\end{equation}
where $\texttt{SelfAttn}(\cdot)$ is the self-attention operation, $\texttt{FgroundAttn}(\cdot)$ is the foreground aware attention operation with 3 layers of mask attention, and $F_{pd}$ is the feature at the middle layer of pixel decoder in SAM.
Similar with the first stage, we predict a refined segmentation mask $P_f\in \mathbb{R}^{H \times W}$ by employing a linear layer to the final features with $F_o$ and the last layer of pixel decoder feature.

The training loss consists of two components, \emph{i.e.}, pixel-wise classification loss and binary mask loss: $\mathcal{L} = \lambda_{ce} \mathcal{L}_{ce} + \lambda_{dice} \mathcal{L}_{dice}$. The total loss of the two-stage spatial-spectral decoder is computed as a weighted sum of the losses from both stages, \emph{i.e.}, $\mathcal{L}_{\text{total}} = \lambda_w \mathcal{L}_{\text{stage1}} + (1 - \lambda_w) \mathcal{L}_{\text{stage2}}$. Here, $\{\lambda_{ce},\lambda_{dice},\lambda_{w}\}$ are weight values:
$\lambda_{ce}=0.25, \lambda_{dice}=0.75, \lambda_{w}=0.8$.

\section{Experiments \& Results}
\subsection{Experimental Setup}
\noindent\textbf{Datasets.}
To evaluate the model performance of our proposed method, we conduct all experiments on a private Gastric Poorly Cohesive Carcinoma (GPCC) dataset and a public Multi-Dimensional Choledoch (MDC) dataset.
The \textbf{Multi-Dimensional Choledoch (MDC) Dataset}~\cite{zhang2019multidimensional} is a publicly available dataset used for hyperspectral image segmentation tasks.
538 scenes are analyzed in this paper, each with 60 spectral bands from $450 nm$ to $750 nm$ and spatial resolution is resized to $256\times 256$ pixels. The dataset focuses on the task of binary Microscopic Hyperspectral Image (MHSI) segmentation.
For experimentation purposes, the MDC dataset is divided into training, validation, and test subsets using a patient-centric hard split method, maintaining a ratio of 3:1:1 for these subsets.
The \textbf{Gastric Poorly Cohesive Carcinoma (GPCC) Dataset} is a private dataset containing 600 scenes with 40 spectral bands. The spatial resolution of each scene is resized to $256\times 256$ pixels. This dataset focuses on a different set of medical imaging challenges, specifically targeting gastric cancer tissues. Similar to the MDC dataset, the GPCC dataset is also partitioned into training, validation, and test subsets using a patient-centric hard split method, maintaining a ratio of 3:1:1 for these subsets.

\begin{table}[t]
\centering
\caption{
Evaluation results of the proposed method and SOTA methods on both MDC and GPCC datasets. The optimal metric values are highlighted in bold.
}
\begin{tabular}{>{\centering\arraybackslash}m{2.7cm}>{\centering\arraybackslash}m{1cm}>{\centering\arraybackslash}m{1cm}>{\centering\arraybackslash}m{1cm}>{\centering\arraybackslash}m{1.3cm}
>{\centering\arraybackslash}m{1cm}>{\centering\arraybackslash}m{1cm}>{\centering\arraybackslash}m{1cm}>{\centering\arraybackslash}m{1.3cm}}
\hline
\multirow{2}{*}{Method} & \multicolumn{4}{c}{MDC} & \multicolumn{4}{c}{GPCC} \\ 
\cmidrule(r){2-5} \cmidrule(r){6-9}
& DSC$\uparrow$ & IoU$\uparrow$ & HD$\downarrow$ &p-value $\downarrow$ & DSC$\uparrow$ & IoU$\uparrow$ & HD$\downarrow$ &p-value $\downarrow$ \\ \hline
3DUNet \cite{cicek2016unet}              &72.48&58.92&73.99&3.06  &70.76&53.79&92.26&3.05 \\ 
BYOL \cite{grill2020bootstrap}          &72.83&60.18&68.54&3.19  &71.73&55.64&86.70&3.15 \\ 
DMVL \cite{liu2020deep}                  &71.97&59.49&72.75&2.96  &70.36&54.37&90.07&3.00 \\ 
HyperNet \cite{wang2021identification}   &72.31&59.54&75.36&3.13  &69.52&54.83&92.99&3.07 \\
SimSiam \cite{chen2021exploring}         &73.42&60.78&68.61&3.25  &71.34&55.48&85.89&3.25 \\ 
SpectralFormer \cite{hong2021spectralformer} &82.68&66.75&56.27&3.76 &79.47&64.57&74.63&3.61 \\
nnUNet \cite{isensee2021nnunet}          &74.06&61.33&71.69&3.58  &71.87&55.32&86.37&3.54 \\ 
Swin-UNETR \cite{tang2022self}           &72.04&58.69&72.80&3.30  &70.93&54.93&89.66&3.11 \\ 
FSS \cite{yun2023factor}                 &74.62&61.61&68.67&3.64  &71.45&54.99&87.48&3.68 \\ 
Spec-Tr \cite{yun2023spectral}         &73.06&60.36&67.96&3.22  &71.60&55.01&86.84&3.27 \\ 
DF-S$^3$R \cite{xie2023exploring}        &75.38&63.83&69.09&3.95  &72.29&56.52&86.13&3.93 \\ 
QSQL-FL \cite{dong2024multi}             &78.39&65.78&63.01&4.24  &75.76&60.44&80.92&4.26 \\ 
VM-UNet \cite{ruan2024vm}          &81.53&67.23&59.75&4.63  &77.53&59.54&77.12&4.66\\
Ours             &\textbf{84.12}&\textbf{69.39}&\textbf{54.84}&-  &\textbf{79.78}&\textbf{66.84}&\textbf{72.96}&-  \\ \hline
\end{tabular}
\label{tab: sota}
\end{table}

The proposed Omni-Fuse extracts primary feature utilizing Swin-Transformer encoder~\cite{liu2021swin} for spatial and MDN~\cite{lin2025mdn} for spectral, and select SAM decoder~\cite{kirillov2023segment} as the decoder.
Data augmentation is employed in Omni-Fuse, combining elastic deformation, rotation and scaling.
The batch size is set to 8 and the maximal training epoch to 300.
AdamW optimizer and weight decay scheduler are utilized with initial learning rate of 0.005.
The training loss is a combination of Cross-Entropy loss and Dice loss.
We select 3 metrics, \emph{i.e.}, Dice Similarity Coefficient (DSC) [\%], Intersection over Union (IoU) [\%], and Hausdorff Distance (HD) [px], for the assessment of model performance.
All experiments are implemented using PyTorch 2.2.0 on an NVIDIA Geforce RTX 3090 GPU.

\begin{figure}[t]
    \centering
    \vspace{-0.4em}
    \includegraphics[width=\linewidth]{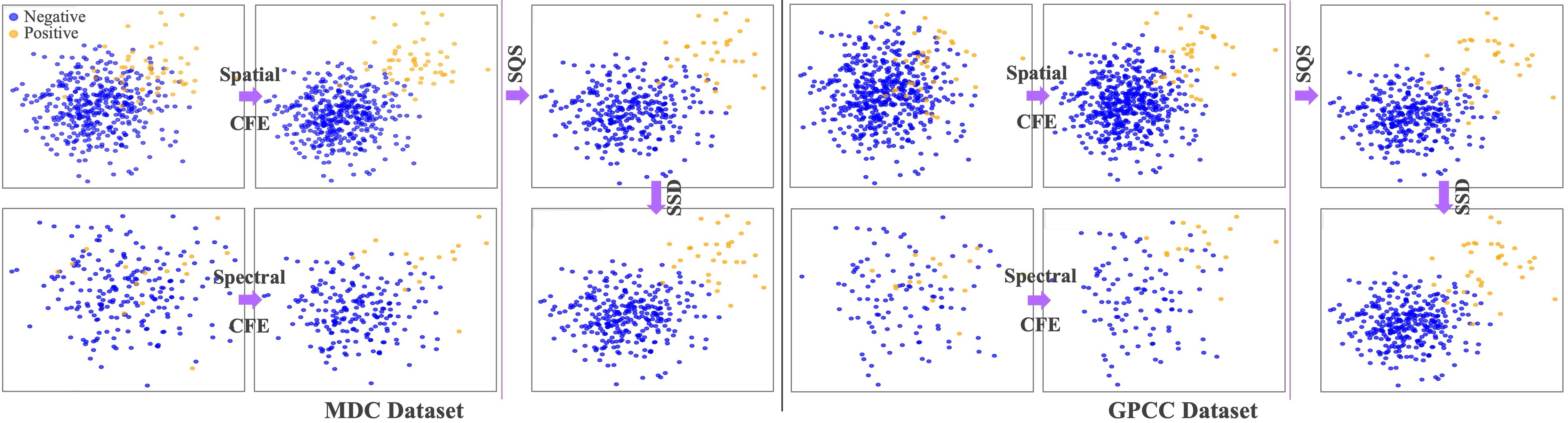}
    \caption{tSNE visualization of the spatial-spectral features processed by several feature fusion modules. ``CFE'' is cross-dimensional feature enhancer, ``SQS'' is spectral-guided spatial query selector, and ``SSD'' is spatial-spectral decoder.}
    \label{fig:tsne}
\end{figure}

\subsection{Performance Evaluation}
We compare segmentation results of the proposed Omni-Fuse with state-of-the-art (SOTA) methods on both MDC and GPCC datasets.
The experimental results in Table~\ref{tab: sota} show that our proposed method outperforms all SOTA methods in all evaluation metrics.
The proposed Omni-Fuse method achieves the highest DSC values, outperforming the second best approach, QSQL-FLs, by a significant margin of 5.73\% on the MDC dataset and 4.02\% on the GPCC dataset.
It also achieves improvements in IoU and HD across both datasets.
And statistical significance tests on DSC metrics show that all p-values below 0.05 on both datasets, indicating that our model's improvements over existing SOTA methods are statistically significant.
All these notable improvements show the segmentation capability of the proposed cross-dimensional feature fusion framework.

More intuitively, we provide tSNE visualization of the iterative features updated by the abundant multi-dimensional fusion blocks in Fig.~\ref{fig:tsne}. 
It is obvious that cross-dimensional feature enhancer can significantly increase the inter-class distance and decrease intra-class distance in both spatial and spectral dimensions.
And this block effectively reduces the spectral redundancy from 0.5755 to 0.46328 on MDC and 0.6450 to 0.5154 on GPCC, indicating its ability to suppress redundant information across adjacent spectral bands.
Spectral-guided spatial query selector and spatial-spectral decoder can help incrementally differentiate cancer area from the background.
Although several attentions are employed for feature fusion, Omni-Fuse maintains low computational complexity and operates at high speed, as the spatial-spectral features are refined progressively.

\begin{table}[t]
\centering
\caption{Evaluation results of the proposed method with various combinations of its modules. ``CNN'' is CNN-based spatial feature extraction, ``Mamba'' is mamba-based spectral feature extraction, ``CFE'' is cross-dimensional feature enhancer, ``SQS'' is spectral-guided spatial query selctor, ``SSD'' os spatial-spectral decoder and ``MR'' is mask refiner. The optimal metric values are highlighted in bold.}
\begin{tabular}{>{\centering\arraybackslash}m{0.8cm}>{\centering\arraybackslash}m{1cm}>{\centering\arraybackslash}m{0.8cm}>{\centering\arraybackslash}m{0.8cm}>{\centering\arraybackslash}m{0.8cm}>{\centering\arraybackslash}m{0.8cm}>{\centering\arraybackslash}m{1cm}>
{\centering\arraybackslash}m{1cm}>{\centering\arraybackslash}m{1cm}>{\centering\arraybackslash}m{1cm}>{\centering\arraybackslash}m{1cm}>{\centering\arraybackslash}m{1cm}>{\centering\arraybackslash}m{1cm}}
\hline
\multirow{2}{*}{CNN} &\multirow{2}{*}{Mamba} &\multirow{2}{*}{CFE} &\multirow{2}{*}{SQS} 
 &\multirow{2}{*}{SSD} &\multirow{2}{*}{MR} &\multicolumn{3}{c}{MDC} &\multicolumn{3}{c}{GPCC}\\
\cmidrule(r){7-9} \cmidrule{10-12}
~ &~ &~ &~ &~ &~ &DSC$\uparrow$ &IoU$\uparrow$ &HD$\downarrow$ &DSC$\uparrow$ &IoU$\uparrow$ &HD$\downarrow$\\ \hline
$\times$  & $\times$ & $\times$ & $\times$ & $\times$ & $\times$ & 70.54 &63.54 &59.67 &67.43 &62.47 &76.97\\ 
$\checkmark$      & $\times$ & $\times$ & $\times$& $\times$ & $\times$& 72.68 &65.75 &58.34 &70.89 &63.28 &75.76\\ 
$\checkmark$      & $\checkmark$ & $\times$ & $\times$ & $\times$ & $\times$ & 76.61 &66.95 &57.23 &74.65 &64.44 &74.32\\ 
$\checkmark$      & $\checkmark$ & $\checkmark$ & $\times$ & $\times$ & $\times$ & 79.39 &68.64 &55.36 &77.23 &65.83 &73.26\\ 
$\checkmark$ &$\checkmark$ &$\checkmark$ &$\checkmark$ &$\times$ &$\times$ &82.23 &68.95 &54.98 &77.95 &65.43 &74.32\\ 
$\checkmark$ &$\checkmark$ &$\checkmark$ &$\checkmark$ &$\checkmark$ &$\times$ &83.47 &68.46 &54.49 &78.47 &66.43 &72.85\\ 
$\checkmark$ &$\checkmark$ &$\checkmark$ &$\checkmark$ &$\checkmark$ &$\checkmark$ & \textbf{84.12}  &\textbf{69.39} &\textbf{54.84} &\textbf{79.78} &\textbf{66.84} &\textbf{72.96} \\ \hline
\end{tabular}
\label{tab:ablation}
\end{table}

\subsection{Ablation Study}
We conduct an ablation study to evaluate each component in Omni-Fuse as shown in Table~\ref{tab:ablation}. The baseline is conducted by removing the CNN, Mamba block, cross-dimension feature enhancer, and two-stage spatial-spectral decoder in the proposed Omni-Fuse. With CNN and Mamba blocks extracting primary spatial and spectral features, DSC has a significant increase of 6.07\%.
Additionally, applying feature enhancement module yields a 2.78\% improvement in DSC on MDC and 2.58\% on GPCC, proving its inter-class discriminative ability.
On top of this, the following spatial query selector, spatial-spectral decoder and mask refinement all make contributions for more accurate tumor area segmentation.
From Table~\ref{tab:ablation}, if the proposed two-stage spatial-spectral decoder (SSD $+$ MR) is replaced with SAM decoder, with enhanced spatial and spectral features added as input, the model performance drops from 84.12\% to 79.39\% on MDC dataset.
Finally, we explore Omni-Fuse using pseudo-color image (420nm, 495nm, and 625nm mapped to RGB) to represent pure spatial information. DSC value decreases 9.74\% on MDC and 8.21\% on GPCC, further highlighting the superiority of hyperspectral image and the proposed omni fusion strategy.

\section{Conclusion}
In this paper, we introduce an omni-fusion strategy of spatial and spectral dimensions for microscopic hyperspectral image segmentation, including a novel cross-dimensional enhancer, a spectral-guided spatial query selector and a two-stage cross-dimensional decoder.
All the deigned modules progressively reinforce the communication between spatial and spectral dimensions, increasing the inter-class distance and decreasing the intra-class distance.
Experiments show our proposed Omni-Fuse achieves much better segmentation performance compared with the SOTA methods with 5.73\% improvement in DSC on MDC dataset and 4.02\% on GPCC dataset.

\begin{credits}
\subsubsection{\ackname} This work was supported by the National Natural Science Foundation of China (Grant No. 62471182), {Shanghai Rising-Star Program (Grant No. 24QA2702100)}, and the Science and Technology Commission of Shanghai Municipality (Grant No. 22DZ2229004).

\subsubsection{\discintname}
The authors have no competing interests in the paper as required by the publisher.
\end{credits}

\bibliographystyle{splncs04}
\bibliography{main}

\begin{thebibliography}{10}
\providecommand{\url}[1]{\texttt{#1}}
\providecommand{\urlprefix}{URL }
\providecommand{\doi}[1]{https://doi.org/#1}

\bibitem{banu2024hyperspectral}
Banu, K.S., Lerma, M., Ahmed, S.U., Gardea-Torresdey, J.L.: Hyperspectral microscopy-applications of hyperspectral imaging techniques in different fields of science: a review of recent advances. Applied Spectroscopy Reviews  \textbf{59}(7),  935--958 (2024)

\bibitem{bengs2020spectral}
Bengs, M., Gessert, N., Laffers, W., Eggert, D., Westermann, S., Mueller, N.A., Gerstner, A.O., Betz, C., Schlaefer, A.: Spectral-spatial recurrent-convolutional networks for in-vivo hyperspectral tumor type classification. In: Proceedings of Medical Image Computing and Computer-Assisted Intervention. pp. 690--699. Springer (2020)

\bibitem{chen2021exploring}
Chen, X., He, K.: Exploring simple siamese representation learning. In: Proceedings of the IEEE/CVF Conference on Computer Vision and Pattern Recognition. pp. 15750--15758 (2021)

\bibitem{cheng2021maskformer}
Cheng, B., Schwing, A.G., Kirillov, A.: Per-pixel classification is not all you need for semantic segmentation (2021)

\bibitem{cicek2016unet}
{\c{C}}i{\c{c}}ek, {\"O}., Abdulkadir, A., Lienkamp, S.S., Brox, T., Ronneberger, O.: 3d u-net: learning dense volumetric segmentation from sparse annotation. In: Proceedings of Medical Image Computing and Computer-Assisted Intervention. pp. 424--432. Springer (2016)

\bibitem{dong2024multi}
Dong, H., Zhou, R., Yun, B., Zhou, H., Zhang, B., Li, Q., Wang, Y.: Multi-stage multi-granularity focus-tuned learning paradigm for medical hsi segmentation. In: Proceedings of Medical Image Computing and Computer-Assisted Intervention. pp. 456--466. Springer (2024)

\bibitem{grill2020bootstrap}
Grill, J.B., Strub, F., Altsché, F., Tallec, F., Richemond, C., Buchatskaya, P., Doersch, E., Pires, C.A., Guo, B., Azar, M.G.: Bootstrap your own latent-a new approach to self-supervised learning. Advances in Neural Information Processing Systems  \textbf{33},  21271--21284 (2020)

\bibitem{hong2021spectralformer}
Hong, D., Han, Z., Yao, J., Gao, L., Zhang, B., Plaza, A., Chanussot, J.: Spectralformer: Rethinking hyperspectral image classification with transformers. IEEE Transactions on Geoscience and Remote Sensing  \textbf{60},  1--15 (2021)

\bibitem{isensee2021nnunet}
Isensee, F., Jaeger, P.F., Kohl, S.A., Petersen, J., Maier-Hein, K.H.: nnu-net: A self-configuring method for deep learning-based biomedical image segmentation. Nature Methods  \textbf{18(2)},  203--211 (2021)

\bibitem{kirillov2023segment}
Kirillov, A., Mintun, E., Ravi, N., Mao, H., Rolland, C., Gustafson, L., Xiao, T., Whitehead, S., Berg, A.C., Lo, W.Y., et~al.: Segment anything. In: Proceedings of the IEEE/CVF international conference on computer vision. pp. 4015--4026 (2023)

\bibitem{li2022asymmetric}
Li, W., Gao, Y., Zhang, M., Tao, R., Du, Q.: Asymmetric feature fusion network for hyperspectral and sar image classification. IEEE Transactions on Neural Networks and Learning Systems  \textbf{34}(10),  8057--8070 (2022)

\bibitem{lin2025mdn}
Lin, S., Yun, B., Shen, W., Li, Q., Yang, A., Wang, Y.: Mdn: Mamba-driven dualstream network for medical hyperspectral image segmentation. In: ICASSP 2025-2025 IEEE International Conference on Acoustics, Speech and Signal Processing (ICASSP). pp.~1--5. IEEE (2025)

\bibitem{liu2020deep}
Liu, B., Yu, A., Yu, X., Wang, R., Gao, K., Guo, W.: Deep multiview learning for hyperspectral image classification. vol. 59(9), pp. 7758--7772 (2020)

\bibitem{liu2021swin}
Liu, Z., Lin, Y., Cao, Y., Hu, H., Wei, Y., Zhang, Z., Lin, S., Guo, B.: Swin transformer: Hierarchical vision transformer using shifted windows. In: Proceedings of the IEEE/CVF international conference on computer vision. pp. 10012--10022 (2021)

\bibitem{ruan2024vm}
Ruan, J., Li, J., Xiang, S.: Vm-unet: Vision mamba unet for medical image segmentation. arXiv preprint arXiv:2402.02491  (2024)

\bibitem{song2024interactive}
Song, L., Feng, Z., Yang, S., Zhang, X., Jiao, L.: Interactive spectral-spatial transformer for hyperspectral image classification. IEEE Transactions on Circuits and Systems for Video Technology  (2024)

\bibitem{tang2022self}
Tang, Y., Yang, Y., Li, D., Roth, H.R., Landman, H., Xu, B., Nath, V., Hatamizadeh, V.: Self-supervised pre-training of swin transformers for 3d medical image analysis. In: Proceedings of the IEEE/CVF Conference on Computer Vision and Pattern Recognition. pp. 20730--20740 (2022)

\bibitem{wang2021identification}
Wang, Q., Sun, L., Wang, X., Zhou, Y., Hu, M., Chen, M., Wen, J., Li, Y., Li, Q.: Identification of melanoma from hyperspectral pathology image using 3d convolutional networks. IEEE Transactions on Medical Imaging  \textbf{40(1)},  218--227 (2021)

\bibitem{xie2023exploring}
Xie, X., Jin, T., Yun, B., Li, B., Wang, Q.: Exploring hyperspectral histopathology image segmentation from a deformable perspective. In: Proceedings of the 31st ACM International Conference on Multimedia. pp. 242--251 (2023)

\bibitem{yao2024specat}
Yao, Z., Liu, S., Yuan, X., Fang, L.: Specat: Spatial-spectral cumulative-attention transformer for high-resolution hyperspectral image reconstruction. In: Proceedings of the IEEE/CVF Conference on Computer Vision and Pattern Recognition. pp. 25368--25377 (2024)

\bibitem{yun2023spectral}
Yun, B., Lei, B., Chen, J., Wang, J., Qiu, H., Shen, S., Li, W., Wang, Q.: Spectral transformer for microscopic hyperspectral pathology image segmentation. IEEE Transactions on Circuits and Systems for Video Technology  (2023)

\bibitem{yun2023factor}
Yun, B., Li, Q., Mitrofanova, Q., Zhou, C., Wang, Y.: Factor space and spectrum for medical hyperspectral image segmentation. In: Proceedings of Medical Image Computing and Computer-Assisted Intervention. pp. 152--162. Springer (2023)

\bibitem{zhang2023hyperspectral}
Zhang, J., Zhao, L., Jiang, H., Shen, S., Wang, J., Zhang, P., Zhang, W., Wang, L.: Hyperspectral image classification based on dense pyramidal convolution and multi-feature fusion. Remote Sensing  \textbf{15}(12), ~2990 (2023)

\bibitem{zhang2019multidimensional}
Zhang, Q., Li, Q., Yu, G., Sun, L., Zhou, M., Chu, J.: A multidimensional choledoch database and benchmarks for cholangiocarcinoma diagnosis. IEEE Access  \textbf{7},  149414--149421 (2019)

\end{thebibliography}

\end{document}